\def\year{2022}\relax

\documentclass[letterpaper]{article} 
\usepackage{aaai22} 
\usepackage{times} 
\usepackage{helvet} 
\usepackage{courier} 
\usepackage[hyphens]{url} 
\usepackage{graphicx} 
\usepackage{natbib} 
\usepackage{caption} 
\DeclareCaptionStyle{ruled}{labelfont=normalfont,labelsep=colon,strut=off} 
\usepackage{algorithm}
\usepackage{algorithmic}

\usepackage[subrefformat=parens]{subcaption}
\usepackage{multirow}
\usepackage{bm}
\usepackage{amsfonts}
\newcommand{\argmin}{\mathop{\rm arg~min}\limits}
\usepackage{amsmath}
\usepackage{color}

\usepackage{newfloat}
\usepackage{listings}
\floatstyle{ruled}
\newfloat{listing}{tb}{lst}{}
\floatname{listing}{Listing}
\title{Audio Visual Scene-Aware Dialog Generation with\\Transformer-based Video Representations}
\author{
  Yoshihiro Yamazaki,
  Shota Orihashi,
  Ryo Masumura,
  Mihiro Uchida,
  Akihiko Takashima}
\affiliations{
  NTT Computer and Data Science Laboratories, NTT Corporation, Japan\\
  yoshihiro.yamazaki.df@hco.ntt.co.jp
}

\usepackage{bibentry}

\begin{document}

\maketitle

\begin{abstract}
There have been many attempts to build multimodal dialog systems that can respond to a question about given audio-visual information, and the representative task for such systems is the Audio Visual Scene-Aware Dialog (AVSD).
Most conventional AVSD models adopt the Convolutional Neural Network (CNN)-based video feature extractor to understand visual information.
While a CNN tends to obtain both temporally and spatially local information, global information is also crucial for boosting video understanding because AVSD requires long-term temporal visual dependency and whole visual information.
In this study, we apply the Transformer-based video feature that can capture both temporally and spatially global representations more efficiently than the CNN-based feature.
Our AVSD model with its Transformer-based feature attains higher objective performance scores for answer generation.
In addition, our model achieves a subjective score close to that of human answers in DSTC10.
We observed that the Transformer-based visual feature is beneficial for the AVSD task because our model tends to correctly answer the questions that need a temporally and spatially broad range of visual information.

\end{abstract}

\begin{figure*}[t]
 \centering
 \includegraphics[width=\textwidth]{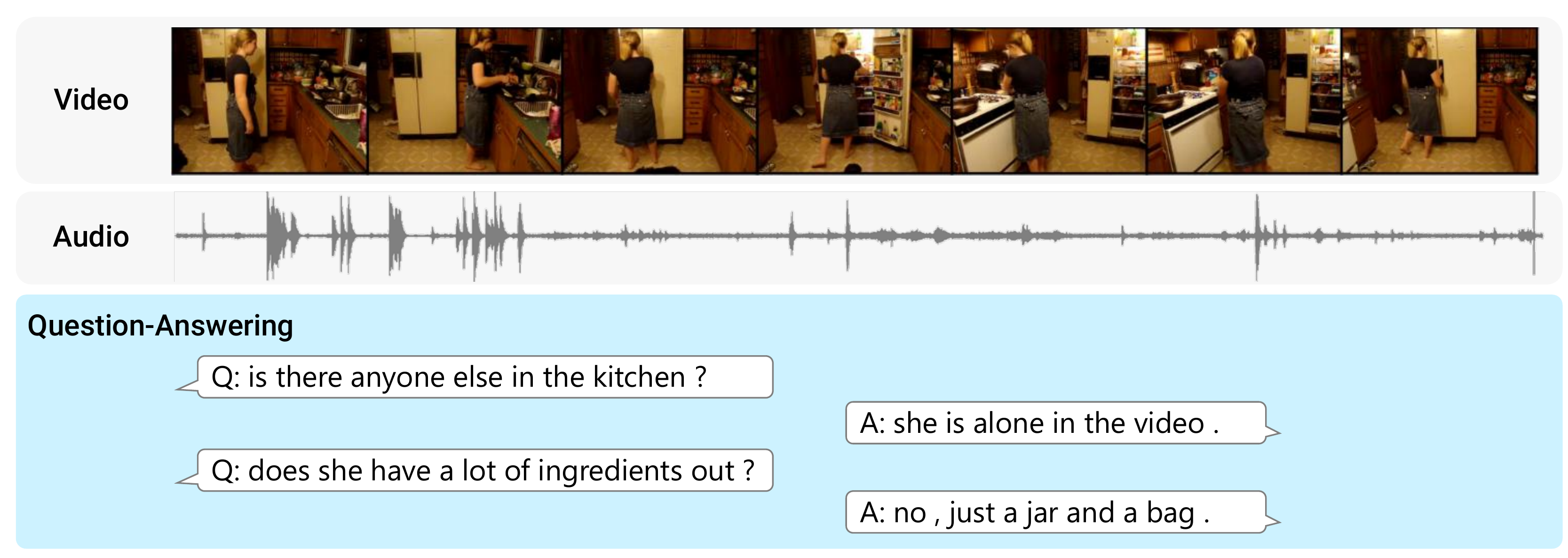}
 \caption{Overview of the Audio Visual Scene-Aware Dialog}
 \label{fig:avsd_explain}
\end{figure*}

\section{Introduction}
While many researchers have recently tackled the text-based conversational agent~\cite{Heck2020trippy,Xu2021beyond}, multimodal dialog systems that incorporate text, audio, and visual cues to determine the response have also been getting attention.
One of the advantages of using multimodal information is that the systems are able to consider more diverse interactions (e.g., dialog systems talking with the user about the events happening around them).
For example, Audio Visual Scene-Aware Dialog (AVSD) has been proposed as the task of multi-turn question-answering based on given text, audio, and video signals~\cite{Nguyen2018from,Hori2019joint,Li2021bridging}.
Figure~\ref{fig:avsd_explain} shows the overview of the AVSD task.
The contents of the audio and video are daily indoor activities, and the dialog system is expected to correctly respond to questions about them.

The AVSD task has been adopted as the competition track in the Dialog System Technology Challenge (DSTC) three times in DSTC7, DSTC8, and DSTC10.
The participants of the AVSD track on the DSTC build their own conversational model by using training data distributed by the organizers, then infer the answers for the test data.
The participants compete with each other based on the objective and subjective performance scores for predicted answers. 
In this paper, we aim to build a model for the AVSD track in DSTC10.

Many of the conventional studies employ neural-based autoregressive text generation for the AVSD model.
Those models encode the text, audio, and video information into latent representations and generate response sentences.
In the previous competition of DSTC8, \citet{Li2021bridging} presented fine-tuning of a pre-trained Transformer-based language model; it showed remarkable performance.
They indicated that pre-training of text generation is beneficial for AVSD, but the quality of visual understanding remains an issue.

Furthermore, the captions and summaries concerning the events involving audio and video are annotated to the dataset of the AVSD, and the participants of the competitions were allowed to use them as input features in DSTC7 and DSTC8.
However, in actual usage of dialog systems, the captions do not exist before the conversation.
In DSTC10, the participants are not allowed to use the captions for inference, only for training.
The captions and summaries are powerful clues in understanding the events because they are concrete and concise representations of the video scenes.
Therefore, in DSTC10, it is necessary to develop a more complete visual understanding than the conventional methods.

Most conventional methods focused on the network architecture in modality fusion or response generation module and relatively little on the video features.
The typical video features for AVSD were extracted from the I3D~\cite{Carreira2017quo} based on 3D-CNN.
CNN-based visual feature extractors tend to output more local information than global information~\cite{Raghu2021vision}.
The conventional AVSD model using a CNN-based video representation had difficulty in answering correctly to questions that need a temporally and spatially broad range of spatio-temporal information.
Thus, promoting the ability to capture that information also seems to be necessary to develop correct answers.
In recent years, TimeSformer~\cite{Bertasius2021is}, a Transformer-based video feature extractor, showed better performance than CNN-based models in action recognition tasks owing to its abilities such as capturing global representations.
Therefore, TimeSformer appears to be effective in developing an AVSD model that can correctly understand visual information and precisely generate answers.

In this paper, we propose to apply TimeSformer-based video features to Transformer-based autoregressive response generation model to enhance visual understanding without recourse to captions of the dialog.
We utilize a pre-trained TimeSformer as a video feature extractor and compare its performances with that of I3D.
Experiments show the improvements in the objective scores.
One of the reason for the results is the ability of TimeSformer to capture a temporally and spatially broad range of visual information.

\section{Related Studies}

\subsection{Network Architectures of AVSD Models}
Almost all AVSD models proposed to date are based on end-to-end neural text generation.
In DSTC7, the encoder-decoder model based on an RNN and an attention module was often adopted~\cite{Nguyen2018from,Hori2019end}.
Many of the models for DSTC7 contained the attention mechanism for multimodal feature fusion and generated natural sentences to some extent.
Transformer-based generation models appeared in DSTC8.
In particular, the pre-trained Transformer-based language model such as GPT-2~\cite{Radford2019language} or BERT~\cite{Devlin2018bert} that was fine-tuned received the top two human-rated scores~\cite{Li2021bridging,Chen2020pretraining}.
Therefore, we employ pre-training and fine-tuning of the Transformer-based language model for our response generation model as it can generate fluent sentence.


\begin{figure*}[t]
 \centering
 \includegraphics[width=\textwidth]{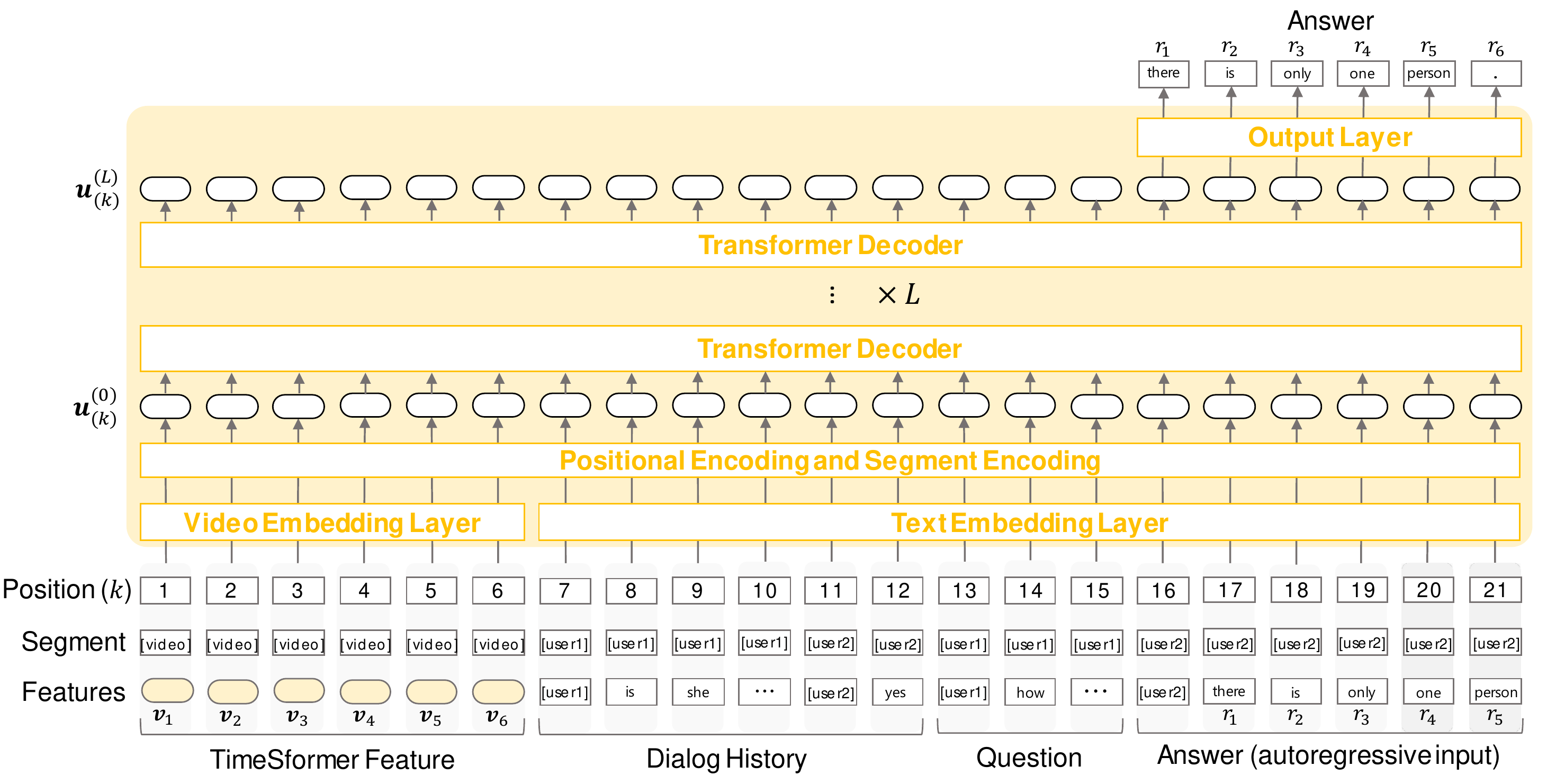}
 \caption{Our response generation model using TimeSformer video feature}
 \label{fig:gpt2_overview}
\end{figure*}

\begin{figure*}[t]
 \centering
 \includegraphics[width=\textwidth]{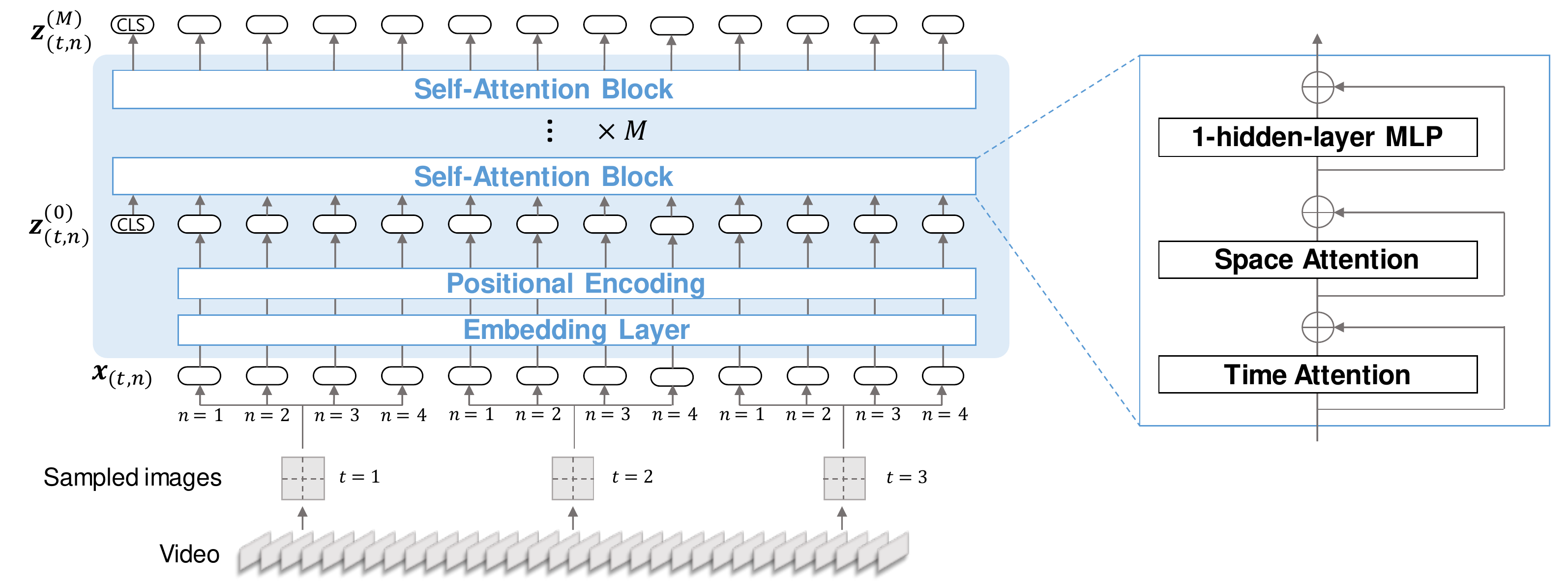}
 \caption{TimeSformer~\cite{Bertasius2021is}, a video feature extractor, with ``Divided Space-Time Attention". The figure illustrates the example in sampling three image frames from a video ($T = 3$) and splitting each frame into four patches ($N = 4$).}
 \label{fig:timesformer}
\end{figure*}

\subsection{Visual Features Used in AVSD}
As mentioned in the previous section, many AVSD attempts tried to improve the network architectures in modality fusion or response generation module, but not video feature extraction.
The video features based on I3D~\cite{Carreira2017quo} were provided by the organizers of the AVSD track, and many studies have used them.
I3D is a visual understanding model based on 3D-CNN; it considers a spatio-temporal relation in converting RGB images or optical flow sequences into visual feature vectors.
On the other hand, there have been some methods that base their AVSD models on other visual features.
\citet{Sanabria2019cmu,Le2020bist,Geng2021dynamic} utilized intermediate representations of ResNet-50~\cite{He2015deep}, ResNeXt-101~\cite{Xie2017aggregated}, and 3D ResNeXt~\cite{Hara2018can}.
Faster-RCNN~\cite{Ren2015faster} has been used to extract region representations for each object~\cite{Le2021vgnmn}.

All the visual features mentioned above are based on CNN.
Transformers can capture temporally and spatially global information more effectively than CNN and so offer better performance.
The state-of-the-art video understanding model based on Transformer, TimeSformer~\cite{Bertasius2021is}, surpasses the CNN-based models in terms of action recognition. 
TimeSformer learns the space-time relationships in the video by using spatial and temporal self-attention.
Thus, it is considered that applying TimeSformer to AVSD leverages the response performance.

\section{AVSD Response Generation Model Using TimeSformer Features}
We propose a Transformer-based autoregressive response generation model using TimeSformer video feature for AVSD.
We extract the video features from a pre-trained TimeSformer model, then train the response generation model.
This section overviews our response generation model, TimeSformer, and its video feature extraction process.

\subsection{Proposed Response Generation Model}
Figure~\ref{fig:gpt2_overview} overviews the proposed Transformer-based response generation model that uses TimeSformer video feature.
The network is based on GPT-2 pre-trained by only text corpora as is done in \citet{Li2021bridging}.
The model takes $I$ frames of video feature $\bm{V} = \{\bm{v}_{(1)}, \cdots , \bm{v}_{(I)}\}$, where $\bm{v}_{(i)}$ is the video feature vector at the $i$-th frame, dialog history $\bm{H}$, and current question $\bm{Q}$ as inputs, and generates $J$ tokens of response sentence $\bm{R} = \{r_{(1)}, \cdots , r_{(J)}\}$, where $r_{(j)}$ is the $j$-th token.
In this paper, ``dialog history" represents the multi-turn question-answer sequence up to the current question $\bm{Q}$.
To decode the response sentence $\bm{R}$, the model predicts the output probability as:
\begin{equation}
  P_{\bm{\Theta}}(\bm{R} \mid \bm{V}, \bm{H}, \bm{Q}) = \prod^{J}_{j=1}P_{\bm{\Theta}}(r_{(j)} \mid \bm{V}, \bm{H}, \bm{Q}, r_{(<j)})
\end{equation}
where $\bm{\Theta}$ denotes a trainable network parameter.

While the conventional method~\cite{Li2021bridging} uses I3D feature as its video feature $\bm{V}$, we use the video feature extracted from the pre-trained TimeSformer model described in the following section.
Input features for the response generation model are the concatenation of TimeSformer video feature, dialog history, and question.

The $k$-th feature embedding $\bm{e}_{\textrm{feat}(k)}$ from the $k$-th TimeSformer and text features are obtained by the video embedding layer and text embedding layer, respectively.
Both of those layers are trainable linear projections.
To capture positional information and explicitly distinguish feature type, the embedding vectors $\bm{e}_{\textrm{pos}(k)}$ and $\bm{e}_{\textrm{segm}(k)}$ are also obtained from the $k$-th positional token and segment token, respectively.
The projection matrix for segment embedding is shared with that of the text embedding.
The input of Transformer decoder blocks $\bm{u}_{(k)}^{(0)}$ is the sum of the feature, position, and segment embedding as follows.
\begin{equation}
  \bm{u}_{(k)}^{(0)} = \bm{e}_{\textrm{feat}(k)} + \bm{e}_{\textrm{pos}(k)} + \bm{e}_{\textrm{segm}(k)}
\end{equation}
Then, $L$ layers of Transformer decoder blocks convert $\bm{u}_{(k)}^{(0)}$ into $\bm{u}_{(k)}^{(L)}$ as follows:
\begin{equation}
  \bm{u}_{(k)}^{(l)} = \textrm{TransformerDecoderBlock}^{(l)}(\bm{u}_{(k)}^{(l-1)})
\end{equation}
where $l$ denotes the layer index of the Transformer decoder blocks.
The output probability for the $j$-th token $r_{(j)}$ is obtained by the output layer which consists of linear projection and the softmax function as follows.
\begin{gather}
  P_{\bm{\Theta}}(r_{(j)} \mid \bm{V}, \bm{H}, \bm{Q}, r_{(<j)}) = \textrm{softmax}(\bm{W}(\bm{u}_{(j+\tau)}^{(L)})) \\
  \tau = I + |\bm{H}| + |\bm{Q}|
\end{gather}
where $\bm{W}$ is a trainable matrix.

The network parameter $\bm{\Theta}$ is optimized to minimize the cross-entropy loss between the output probabilities of predicted and reference tokens as:
\begin{equation}
  \bm{\hat{\Theta}} = \argmin_{\bm{\Theta}}-\log P_{\bm{\Theta}}(\bm{R} \mid \bm{V}, \bm{H}, \bm{Q})
\end{equation}
where $\bm{\hat{\Theta}}$ denotes an optimized network parameter.


\subsection{Architecture of TimeSformer}
\citet{Bertasius2021is} proposed TimeSformer, a video feature extractor that applies spatio-temporal self-attention to sequences of image patches.
They introduced several variants of methods involving self-attention.
In this study, we use ``Divided Space-Time Attention", in which temporal and spatial attention are used separately, because this approach yielded the best performance on action recognition tasks.
Figure~\ref{fig:timesformer} shows the architecture of TimeSformer with ``Divided Space-Time Attention".

Fixed $T$ frames of RGB image sequences are sampled from the raw video.
Each image is split into $N$ patches, each of which has size of $D \times D$ pixels following process used in ViT~\cite{Dosovitskiy2020image}.
Each patch is flatten into a vector $\bm{x}_{(t,n)} \in \mathbb{R}^{3D^{2}}$ where $t$ and $n$ denote frame index and patch index, respectively.
The input feature of the network is the series of the vectors $\{\bm{x}_{(1,1)}, \bm{x}_{(1,2)}, \cdots, \bm{x}_{(2,1)}, \bm{x}_{(2,2)}, \cdots, \bm{x}_{(T,N)}\}$.
Then, $\bm{x}_{(t,n)}$ is linearly embedded into the inner representation, and the positional embedding is added to it to capture the sequential information.
Let $\bm{z}^{(0)}_{(t,n)} \in \mathbb{R}^{d}$ be the obtained vector, where $d$ is the number of units of the hidden layer.
Here, vector $\bm{z}^{(0)}_{(0,0)}$ is taken as the classification (CLS) token appended to the top of the series of $\bm{z}^{(0)}_{(t,n)}$ as in BERT~\cite{Devlin2018bert}.
The sequence of $\bm{z}^{(0)}_{(t,n)}$ passes through $M$ layers of self-attention blocks and results in $\bm{z}^{(M)}_{(t,n)}$.
In a self-attention block, each patch embedding is used for time attention with the patch on the same position across different frames to extract temporal dependency.
Space attention is applied to each patch embedding with all the patches in the same frame to capture spatial dependency followed by the one-hidden-layer of the feed-forward network.
Layer normalization and residual connections are used for each operation.

\subsection{Extraction of TimeSformer feature}
The original TimeSformer model is trained for action recognition, and the CLS token on the last layer $\bm{z}^{(M)}_{(0,0)}$ is used for prediction.
However, it is desired to obtain the feature vectors for each frame to more fully utilize the rich information of the video.
Thus, we used the averaged vector $\bm{\hat{v}}_{(t)}$ along with all patches $\bm{z}^{(M)}_{(t,n)}$ in each frame as the visual feature for AVSD.
\begin{equation}
  \bm{\hat{v}}_{(t)} = \frac{1}{N}\sum_{n=1}^{N}\bm{z}^{(M)}_{(t,n)}
\end{equation}

Note that $T$, the number of frames, is constant number throughout their pre-training phase of TimeSformer. 
Fixed $T$ frames of feature vectors are obtained regardless of the length of the video, where the density of information depends on the length of the video, which is considered to have a negative impact on learning our AVSD model.
Thus, we compare two methods to extract video features $\bm{V}$.
One is to simply extract a fixed number of frames, and the other is to extract a variable number of frames such that all features from the sequence with different durations have equal density.

\begin{figure}[t]
 \centering
 \begin{tabular}{c}
  \begin{minipage}[t]{1.0\hsize}
   \includegraphics[width=\columnwidth]{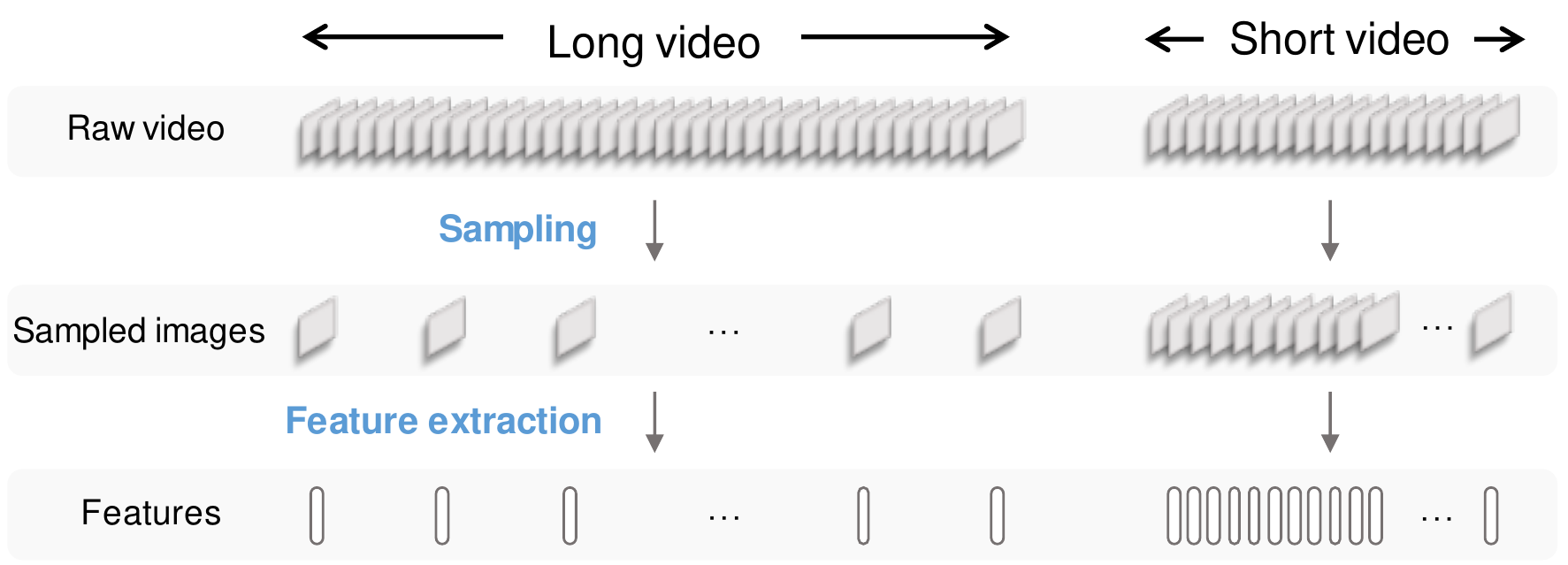}
   \subcaption{Fixed-frame Extraction}
   \label{fig:fixed}
  \end{minipage} \\ \\
  \begin{minipage}[t]{1.0\hsize}
   \includegraphics[width=\columnwidth]{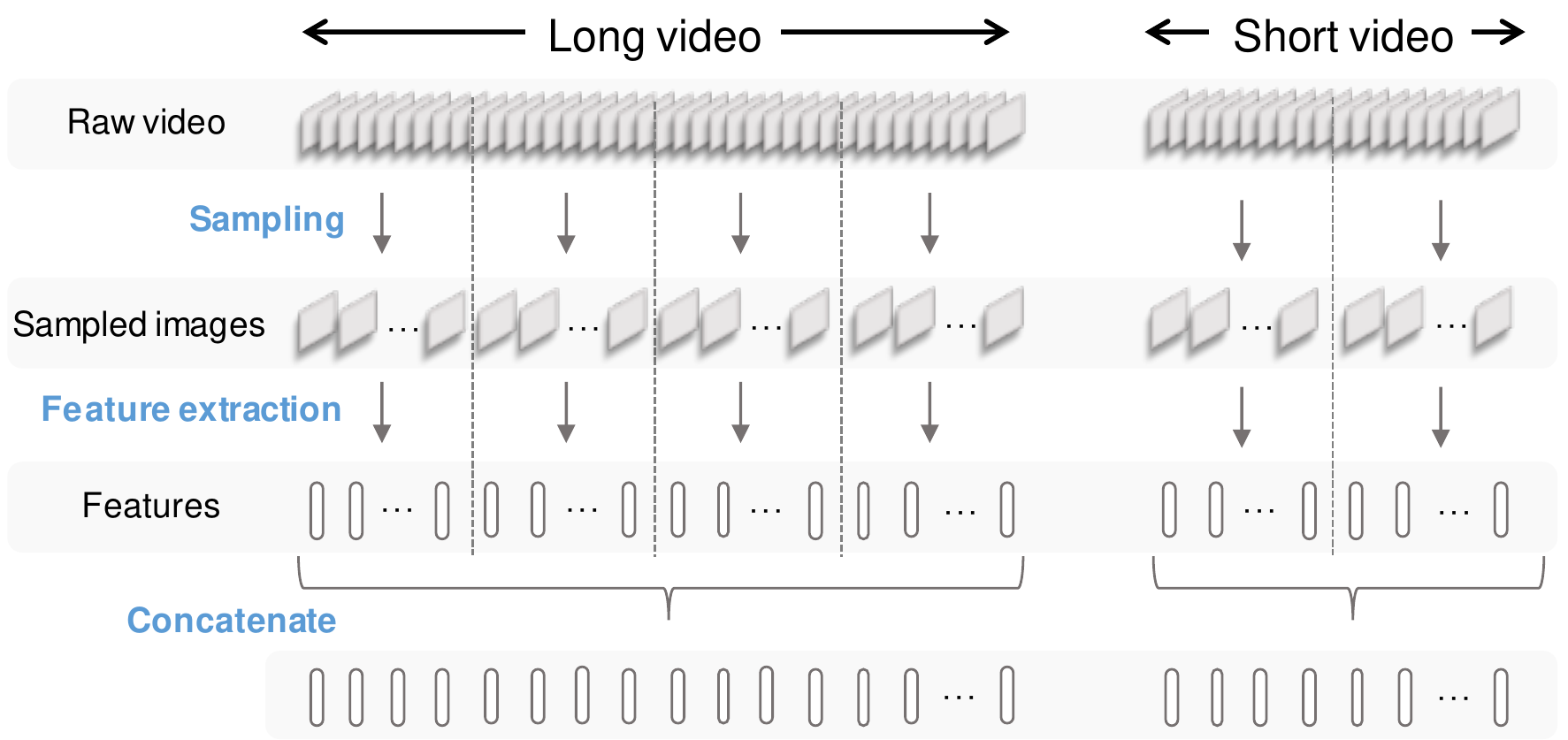}
   \subcaption{Variable-frame Extraction}
   \label{fig:variable}
  \end{minipage}
 \end{tabular}
 \caption{Feature extraction methods using TimeSformer}
\end{figure}

\subsubsection{Fixed-frame Extraction}
First, we consider a simple method to extract the vectors $\bm{V}$ for a fixed number of frames.
Figure~\ref{fig:fixed} shows how to extract the vectors.
The sampling method is to extract fixed $T$ frames from the entire video.
In this method, $\bm{V}$ and $I$, the number of frames of $\bm{V}$, are indicated as follows.
\begin{gather}
    \bm{V} = \{\bm{\hat{v}}_{(1)}, \cdots , \bm{\hat{v}}_{(T)}\} \\ 
    I = T
\end{gather}
Since the number of frames to be extracted remains the same regardless of the length of the video, the frame rate of feature extraction changes for each video.
It is expected that this inconsistency in density degrades model performance.

\subsubsection{Variable-frame Extraction}
On the other hand, we also consider a method to extract vectors for a variable number of frames at equal intervals depending on the length of the video using the pre-trained TimeSformer; the number of input and output frames is assumed to be fixed.
Figure~\ref{fig:variable} shows the vector extraction method.
In order to obtain vectors for a variable number of frames from the pre-trained TimeSformer, we split the video into $S$ segments, extract fixed $T$ frames of feature vectors from each segment, and combine them.
First, the video segments that do not meet the fixed time length are complemented by copies of the last frame.
Then, for each segment, fixed $T$ frames of feature vectors are obtained using the pre-trained model of TimeSformer.
Let $\bm{\hat{v}}_{(s,t)}$ be a feature vector obtained from the $t$-th frame in the $s$-th segment.
Finally, by combining the vectors extracted from all segments and excluding the vectors corresponding to the complemented $T_{\textrm{copy}}$ frames, video representations $\bm{V}$ can be obtained at equal intervals that depend on video length.
In this method, $\bm{V}$ and $I$ are indicated as follows.
\begin{multline}
    \bm{V} = \{\bm{\hat{v}}_{(1,1)}, \cdots , \bm{\hat{v}}_{(1,T)}, \bm{\hat{v}}_{(2,1)}, \cdots ,\\ \bm{\hat{v}}_{(S-1,T)}, \bm{\hat{v}}_{(S,1)}, \cdots , \bm{\hat{v}}_{(S,T-T_{\textrm{copy}})}\}    
\end{multline}
\begin{gather}
    I = ST - T_{\textrm{copy}}
\end{gather}

\section{Experimental Setup}
We trained and evaluated the conventional response generation model that uses the I3D video feature~\cite{Li2021bridging} and our model using TimeSformer video feature by using the test sets of DSTC7 and DSTC8.
We compared two conditions for TimeSformer feature extraction (fixed-frame and variable-frame).

Here, the conventional response generation model utilized not only video but also the audio features from Vggish~\cite{Hershey2017cnn} as non-linguistic information.
Although the use of Vggish improved the objective scores to some extent, the improvement was slight.
Therefore, we did not use the Vggish features with TimeSformer.
To compare the performances under the conditions without the Vggish, we evaluated the model only uses I3D.
Since a model ensemble is effective to improve generalization performance, we also built an ensemble of models of each condition.

We submitted the predicted answers for the test set of the DSTC10 by using our model.
We reported the performances in the DSTC10.

\begin{table*}[t]
  \centering
  \small
  \caption{Objective result from the test set of DSTC7}
  \label{tab:dstc7_result}
  \begin{tabular}{l|ccccccc}\hline
    \multicolumn{1}{c|}{Conditions} & BLEU-1 & BLEU-2 & BLEU-3 & BLEU-4 & METEOR & ROUGE-L & CIDEr \\ \hline
    I3D and Vggish \cite{Li2021bridging} & 0.673 & 0.544 & 0.445 & 0.366 & 0.248 & 0.527 & 0.904 \\
    I3D and Vggish (ensemble) & 0.694 & \textbf{0.573} & \textbf{0.478} & \textbf{0.404} & 0.255 & 0.544 & 1.043 \\
    I3D & 0.670 & 0.542 & 0.440 & 0.359 & 0.246 & 0.526 & 0.929 \\
    I3D (ensemble) & 0.675 & 0.547 & 0.447 & 0.368 & 0.247 & 0.529 & 0.942 \\
    TimeSformer fixed-frame & 0.692 & 0.569 & 0.473 & 0.398 & \textbf{0.256} & 0.546 & 1.044 \\
    TimeSformer fixed-frame (ensemble) & 0.692 & 0.570 & 0.476 & 0.402 & \textbf{0.256} & 0.546 & \textbf{1.057} \\
    TimeSformer variable-frame & 0.691 & 0.567 & 0.471 & 0.397 & 0.255 & 0.543 & 1.048 \\
    TimeSformer variable-frame (ensemble) & \textbf{0.695} & 0.572 & 0.477 & 0.403 & 0.255 & \textbf{0.547} & 1.049 \\ \hline
  \end{tabular}
  
  \centering
  \small
  \caption{Objective result from the test set of DSTC8}
  \label{tab:dstc8_result}
  \begin{tabular}{l|ccccccc}\hline
    \multicolumn{1}{c|}{Conditions} & BLEU-1 & BLEU-2 & BLEU-3 & BLEU-4 & METEOR & ROUGE-L & CIDEr \\ \hline
    I3D and Vggish \cite{Li2021bridging} & 0.670 & 0.548 & 0.452 & 0.378 & 0.245 & 0.539 & 0.979 \\
    I3D and Vggish (ensemble) & 0.675 & 0.557 & 0.465 & 0.392 & 0.248 & 0.542 & 1.018 \\
    I3D & 0.669 & 0.547 & 0.454 & 0.379 & 0.244 & 0.534 & 0.970 \\
    I3D (ensemble) & 0.671 & 0.552 & 0.459 & 0.385 & 0.247 & 0.539 & 0.992 \\
    TimeSformer fixed-frame & 0.679 & 0.559 & 0.466 & 0.392 & \textbf{0.252} & 0.547 & 1.032 \\
    TimeSformer fixed-frame (ensemble) & \textbf{0.682} & \textbf{0.563} & \textbf{0.471} & \textbf{0.398} & 0.251 & \textbf{0.548} & \textbf{1.049} \\
    TimeSformer variable-frame & 0.681 & 0.561 & 0.467 & 0.393 & 0.250 & 0.546 & 1.037 \\
    TimeSformer variable-frame (ensemble) & 0.680 & 0.561 & 0.467 & 0.393 & 0.250 & 0.546 & 1.028 \\ \hline
  \end{tabular}

  \centering
  \small
  \caption{Objective and subjective result from the test set of DSTC10 evaluated by the organizers}
  \label{tab:dstc10_result}
  \begin{tabular}{l|cccccccc}\hline
    \multicolumn{1}{c|}{Conditions} & BLEU-1 & BLEU-2 & BLEU-3 & BLEU-4 & METEOR & ROUGE-L & CIDEr & Human \\ \hline
    DSTC10 Baseline~\cite{Shah2021audio} & 0.572 & 0.422 & 0.320 & 0.247 & 0.191 & 0.439 & 0.566 & 2.851 \\
    TimeSformer fixed-frame (ensemble) & 0.680 & 0.558 & 0.461 & 0.385 & 0.247 & 0.539 & 0.957 & 3.567 \\
    TimeSformer variable-frame (ensemble) & 0.679 & 0.554 & 0.456 & 0.379 & 0.246 & 0.536 & 0.945 & - \\ 
    Ground Truth & - & - & - & - & - & - & - & 3.958 \\ \hline
  \end{tabular}
\end{table*}

\begin{figure*}[t]
 \centering
 \includegraphics[width=\textwidth]{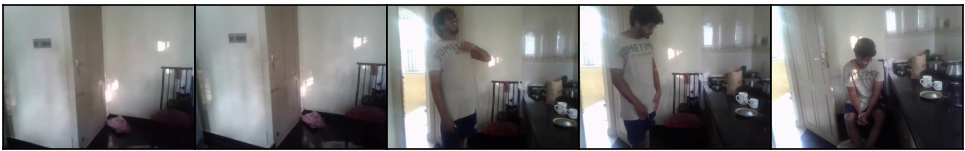}
 \begin{tabular}{lll} \hline
  Question & & how many people are in the room? \\ \hline
  \multirow{9}{*}{Answer} & I3D and Vggish \cite{Li2021bridging} & there are two people in the room.\\
  & I3D and Vggish (ensemble) & there are two people in the room.\\
  & I3D & there are two people in the room\\
  & I3D (ensemble) & there are two people in the room.\\
  & TimeSformer fixed-frame & there is only one person in the room.\\
  & TimeSformer fixed-frame (ensemble) & there is only one person in the room.\\
  & TimeSformer variable-frame & there are two people in the room.\\
  & TimeSformer variable-frame (ensemble) & there are two people in the room.\\
  & Ground Truth & there is just one man \\ \hline
 \end{tabular}
 %
 %
 \caption{A question-answering sample that TimeSformer could answer correctly}
 \label{fig:good_sample}
\end{figure*}

\begin{figure*}[t]
 \centering
 \includegraphics[width=\textwidth]{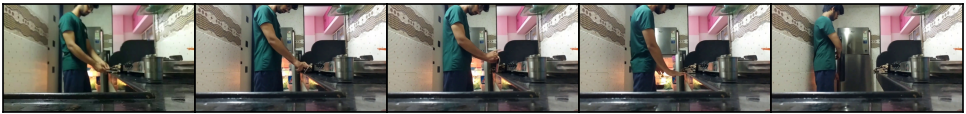}
 \begin{tabular}{lll} \hline
  Question & & is this person wearing glasses? \\ \hline
  \multirow{9}{*}{Answer} & I3D and Vggish \cite{Li2021bridging} & yes, he is wearing glasses.\\
  & I3D and Vggish (ensemble) & yes, he is wearing glasses.\\
  & I3D & yes, he is wearing glasses.\\
  & I3D (ensemble) & yes, he is wearing glasses.\\
  & TimeSformer fixed-frame & no, he is not wearing glasses.\\
  & TimeSformer fixed-frame (ensemble) & no, he is not wearing glasses.\\
  & TimeSformer variable-frame & yes, he is wearing glasses.\\
  & TimeSformer variable-frame (ensemble) & no, he is not wearing glasses.\\
  & Ground Truth & yes he is wearing glasses \\ \hline
 \end{tabular}
 %
 %
 \caption{A question-answering sample that TimeSformer had difficulty in answering correctly}
 \label{fig:bad_sample}
\end{figure*}

\subsection{Datasets}
The AVSD track uses a multimodal dialog corpus holding questions and answers about daily life activities.
The AVSD corpus was constructed by annotating the questions-answers to the videos in the Charades dataset~\cite{Sigurdsson2016hollywood}.
The annotators consist of a questioner and an answerer.
The questioner asked the answerer about the contents of the audio and video material, and the answerer responded to the question posed.
For each video, ten question-answering turns were conducted.
Here, six ground truth answers were annotated by six different answerers in the test set.
The corpus also contains the captions and summaries about the events except for the test set of DSTC10.
The number of videos in the training and validation sets were 7,659 and 1,787, respectively, and that in the test sets was 1,710 in DSTC7 and DSTC8 and 1,804 in DSTC10.


\subsection{Video Feature Extraction}
We used the I3D-flow, I3D-rgb, and Vggish features distributed by the organizers of DSTC10 as the conventional methods.
I3D-flow and I3D-rgb were trained to solve the action recognition task by using the Kinetics dataset. 
The number of dimensions per frame was 2,048.
Vggish was trained to predict the audio class label from Youtube videos.
The number of dimensions per frame was 128.

TimeSformer feature vectors were obtained from the model\footnote{https://github.com/facebookresearch/TimeSformer} pre-trained by using the HowTo100M action recognition dataset~\cite{Miech2019howto100m}.
The model's input consisted of thirty two $224\times224$ image sequences.
Thus, when we extracted the visual features from the pre-trained TimeSformer, we sampled 32 frames from the original videos from the Charades dataset~\cite{Sigurdsson2016hollywood} and resized them to $224\times224$.
Each patch had size of $16\times16$ pixels.
The number of dimensions per frame was 768.

\subsection{Response Generation Model}
We fine-tuned the pre-trained GPT-2 (12-layer, 768-hidden, 12-heads, 117M parameters) released by the Huggingface Transformers~\cite{Wolf2020transformers} for the AVSD dataset.
The text features were tokenized by using WordPieces~\cite{Wu2016googles} as was done in \citet{Li2021bridging}.

The batchsize was four, and the optimization algorithm was AdamW~\cite{Loshchilov2017decoupled} with learning rate of $6.25 \times 10^{-5}$.
The models were trained using four epochs to minimize the cross-entropy loss.
In decoding the answer sentence, we applied beam-search with a beam width of five, a max length of 20, and a length penalty of 0.3.
We only examined beam-search because \citet{Li2021bridging} reported that it is more suitable to AVSD than other decoding methods such as greedy-search and nucleus sampling.
We also built an ensemble of models to improve generalization performance.
We trained the four different models by varying the random seed, averaged the probability distributions of the trained models, and decoded the sentence.

\section{Results}
An objective evaluation was conducted by using the answers generated for the test sets of DSTC7 and DSTC8.
We calculated the metrics based on word overlaps, such as BLEU, METEOR, ROUGE-L, and CIDEr, using the MSCOCO evaluation toolkit\footnote{https://github.com/tylin/coco-caption}.

Tables~\ref{tab:dstc7_result} and \ref{tab:dstc8_result} show the objective evaluation results for the test sets of DSTC7 and DSTC8, respectively.
Note that ``I3D and Vggish~\cite{Li2021bridging}" represents our experimental results, not the scores reported in their paper.
On the whole, TimeSformer-based models were superior to the I3D-based models, though the ensemble of I3D and Vggish was competitive with TimeSformer in the test set of DSTC7. 
This result indicates that TimeSformer is a more suitable visual feature extractor than I3D for AVSD.
TimeSformer fixed-frame and variable-frame performed well equally.
The fixed-frame feature captures the whole information of the video, but it is sparse or dense depending on video duration.
To the contrary, the variable-frame feature is uniformly extracted from the entire video, but it is unlikely to capture the temporal dependency of the whole video because feature extraction is based on segments.
Since the two extraction methods have pros and cons, there was not so many differences between them.
In addition, the ensemble models achieved better scores in almost all conditions.

The answers for the test set of DSTC10 were generated by using the ensemble of TimeSformer fixed-frame and variable-frame, and we submitted the prediction results to the organizers
who conducted objective and subjective evaluations.
The metrics of the objective evaluation are the same as used in the experiments for the test sets of DSTC7 and DSTC8.
The evaluators rated the generated responses considering correctness, naturalness, and informativeness using a five-grade scale (one: very poor, five: very good).

Table~\ref{tab:dstc10_result} shows our competition results for DSTC10.
The table also shows the results of the baseline system by the organizers based on a Transformer encoder-decoder using I3D and Vggish~\cite{Shah2021audio} and the subjective score for the ground truth answers.
The subjective evaluation for our models examined only the fixed-frame model.
Our TimeSformer-based model surpassed the baseline in both subjective and objective scores.
Moreover, our model achieved a close-to-human rating against the ground truth, which indicates the suitability of the Transformer-based video feature to AVSD.

\section{Discussion}
We investigated the tendency of the TimeSformer-based model in order to discern the cause of the improved performance and the remaining challenges.

The TimeSformer-based model correctly answered the question of how many people were shown in the video more often than the I3D-based model (e.g., Figure~\ref{fig:good_sample}).
To grasp the number of people in the video, the model must capture the global spatio-temporal dependency so as to detect people in each frame and track each person across the frames.
The TimeSformer fixed-frame model correctly determined and answered the number of people due to its ability to catch a broad range of temporal relationships.
However, the TimeSformer-based model tended to incorrectly answer the questions that needed local visual information.
As shown in Figure~\ref{fig:bad_sample}, the TimeSformer-based model failed to recognize whether the man was wearing glasses or not. 
To answer this question correctly, the model must pay attention to the man's head, which the CNN-based I3D model is proficient at.
These tendencies suggest that the model should extract local or global features of the video depending on question content for further improvement.

In addition, all models had trouble answering when the video was unclear or viewpoint movement was rapid.
Therefore, stable feature extraction from low-quality or complex movement videos is required.

\section{Conclusion}
In this paper, we proposed to apply the Transformer-based video representations instead of the CNN-based representations to the autoregressive response generation model for AVSD.
The results of a subjective evaluation for the test sets of DSTC7 and DSTC8 showed that the Transformer-based model outperformed the CNN-based model.
Our model was competitive with the ground truth answers for DSTC10.
The Transformer-based model was likely to answer properly the question about the number of people shown in the video; a task that needs the spatio-temporal global dependencies of the video.

In the future, we will construct a model that flexibly extracts local or global visual information depending on the pattern of the question.
In addition, we plan to improve the visual understanding of low-quality and/or complex videos via data expansion.

\bibliography{211027_dstc10,myrefs}


\end{document}